\documentclass{svmult}
\usepackage{makeidx} 
\usepackage{graphicx}
\usepackage{multicol}

\usepackage{amsmath}
\usepackage{amssymb}
\usepackage{bm}

\newcommand{\rar}{\rightarrow}

\newcommand{\ttt}{\texttt}

\newcommand{\mca}{\mathcal}

\makeindex       

\begin{document}

\title*{The Dilated Triple}
\author{Marko A. Rodriguez \and Alberto Pepe \and Joshua Shinavier}

\institute{
	Marko A. Rodriguez \at T-5, Center for Nonlinear Studies, Los Alamos National Laboratory, \email{marko@lanl.gov}
	\and Alberto Pepe \at Center for Embedded Networked Sensing, University of California at Los Angeles, \email{apepe@ucla.edu}
	\and Joshua Shinavier \at Tetherless World Constellation, Rensselaer Polytechnic Institute \email{josh@fortytwo.net}
}

\maketitle

\abstract{The basic unit of meaning on the Semantic Web is the RDF statement, or triple, which combines a distinct subject, predicate and object to make a definite assertion about the world.  A set of triples constitutes a graph, to which they give a collective meaning.  It is upon this simple foundation that the rich, complex knowledge structures of the Semantic Web are built.  Yet the very expressiveness of RDF, by inviting comparison with real-world knowledge, highlights a fundamental shortcoming, in that RDF is limited to statements of absolute fact, independent of the context in which a statement is asserted. This is in stark contrast with the thoroughly context-sensitive nature of human thought. The model presented here provides a particularly simple means of contextualizing an RDF triple by associating it with related statements in the same graph. This approach, in combination with a notion of graph similarity, is sufficient to select only those statements from an RDF graph which are subjectively most relevant to the context of the requesting process.}

\begin{footnotesize}
$\;$
\newline
Rodriguez, M.A., Pepe, A., Shinavier, J., ``The Dilated Triple," Emergent Web Intelligence: Advanced Semantic Technologies, Advanced Information and Knowledge Processing series, eds. R. Chbeir, A. Hassanien,  A. Abraham, and Y. Badr, Springer-Verlag, pages 3-16, ISBN:978-1-84996-076-2, June 2010.
\end{footnotesize}

\section{Introduction}

The World Wide Web introduced a set of standards and protocols that has led to the development of a collectively generated graph of web resources. Individuals participate in creating this graph by contributing digital resources (e.g.~documents, images, etc.) and linking them together by means of dereferenceable Hypertext Transfer Protocol (HTTP) Uniform Resource Identifiers (URI) \cite{lee94}. While the World Wide Web is primarily a technology that came to fruition in the early nineties, much of the inspiration that drove the development of the World Wide Web was developed earlier with such systems as Vannevar Bush's visionary Memex device \cite{vbush} and Ted Nelson's Xanadu \cite{nelsonht}. What the World Wide Web provided that made it excel as the \textit{de facto} standard was a common, relatively simple, distributed platform for the exchange of digital information. The World Wide Web has had such a strong impact on the processes of communication and cognition that it can be regarded as a revolution in the history of human thought -- following those of language, writing and print \cite{harnad1991}.

While the World Wide Web has provided an infrastructure that has revolutionized the way in which many people go about their daily lives, over the years, it has become apparent that there are shortcomings to its design. Many of the standards developed for the World Wide Web lack a mechanism for representing ``meaning'' in a form that can be easily interpreted and used by machines. For instance, the majority of the Web is made up of a vast collection of Hypertext Markup Language (HTML) documents. HTML documents are structured such that a computer can discern the intended layout of the information contained within a document, but the content itself is expressed in natural language and thus, understandable only to humans. Furthermore, all HTML documents link web resources according to a single type of relationship. The meaning of a hypertext relationship can be loosely interpreted as ``cites'' or ``related to''.  The finer, specific meaning of this relationship is not made explicit in the link itself. In many cases, this finer meaning is made explicit within the HTML document. Unfortunately, without sophisticated text analysis algorithms, machines are not privy to the communication medium of humans. Yet, even within the single relationship model, machines have performed quite well in supporting humans as they use go about discovering and sharing information on the World Wide Web  \cite{anatom:brin1998,hits:kleinberg1999,topic:haveliwala2002,tagging:hub2006}.

\begin{quote}
The Web was designed as an information space, with the goal that it should be useful not only for human-human communication, but also that machines would be able to participate and help. One of the major obstacles to this has been the fact that most information on the Web is designed for human consumption, and even if it was derived from a database with well defined meanings (in at least some terms) for its columns, that the structure of the data is not evident to a robot browsing the web. \cite{berners:roadmap1998}
\end{quote}

As a remedy to the aforementioned shortcoming of the World Wide Web, the Semantic Web initiative has introduced a standard data model which makes explicit the type of relationship that exists between two web resources \cite{lee:semantic2001,pubsem:lee2001}. Furthermore, the Linked Data community has not only seen a need to link existing web resources in meaningful ways, but also a need to link the large amounts of non-typical web data (e.g.~database information) \cite{linkeddata:bizer2008}.\footnote{The necessity to expose large amounts of data on the Semantic Web has driven the development of triple-store technology. Advanced triple-store technology parallels relational database technologies by providing an efficient medium for the storage and querying of semantic graphs \cite{lee:triple2004,owlim:kiryakov2005,agraph:aasman2006}.} The standard for relating web resources on the Semantic Web is the Resource Description Framework (RDF) \cite{lee:semantic2001,rdfcon:klyne2004}. RDF is a data model\footnote{RDF is a data model, not a serialization format. There exist various standard serialization formats such as RDF/XML, N3 \cite{n3:lee1998}, Turtle \cite{turtle:beckette2006}, Trix \cite{trix:carroll2004}, etc.} that is used to create graphs of the form
\begin{equation}
	R \subseteq \underbrace{(U \cup B)}_\text{subject} \times \underbrace{U}_\text{predicate} \times \underbrace{(U \cup B \cup L)}_\text{object},
\end{equation}
where $U$ is the infinite set of all URIs \cite{uri:2001,uri:berners2005}, $B$ is the infinite set of all blank nodes, and $L$ is the infinite set of all literal values.\footnote{Other formalisms exist for representing an RDF graph such as the directed labeled graph, bipartite graph \cite{hayes:birdf2004}, and directed hypergraph models \cite{hyperrdf:2006}.} An element in $R$ is known as a statement, or triple, and it is composed of a set of three elements: a subject, a predicate, and an object. A statement in RDF asserts a fact about the world. 

\begin{quote}
``The basic intuition of model-theoretic semantics is that asserting a sentence makes a claim about the world: it is another way of saying that the world is, in fact, so arranged as to be an interpretation which makes the sentence true. In other words, an assertion amounts to stating a constraint on the possible ways the world might be.'' \cite{rdfsem:hayes2004}
\end{quote}
An example RDF statement is $(\ttt{lanl:marko}, \ttt{foaf:knows}, \ttt{ucla:apepe})$.\footnote{All resources in this article have been prefixed in order to shorten their lengthy namespaces. For example, \ttt{foaf:knows}, in its extended form, is \ttt{http://xmlns.com/foaf/0.1/knows}.} This statement makes a claim about the world: namely that ``Marko knows Alberto".  The \ttt{foaf:knows} predicate defines the meaning of the link that connects the subject \ttt{lanl:marko} to the object \ttt{ucla:apepe}. On the World Wide Web, the only way that such semantics could be derived in a computationally efficient manner would be to note that in Marko's webpage there exists an \ttt{href} link to Alberto's webpage. While this web link does not necessarily mean that ``Marko knows Alberto'', it is the simplest means, without text analysis techniques, to recognize that there exists a relationship between Marko and Alberto. Thus, for machines, the World Wide Web is a homogenous world of generic relationships. On the Semantic Web, the world is a rich, complicated network of meaningful relationships.

The evolution from the World Wide Web to the Semantic Web has brought greater meaning and expressiveness to our largest digital information repository \cite{sem:hellman1999}. This explicit meaning provides machines a richer landscape for supporting humans in their information discovery and sharing activities. However, while links are typed on the Semantic Web, the meaning of the type is still primarily based on human interpretation. Granted this meaning is identified by a URI, however, for the machine, there exists no meaning, just symbols that it has been ``hardwired'' to handle \cite{uschold:sem2001}.

\begin{quote}
``Machine usable content presumes that the machine knows what to do with information on the Web.  One way for this to happen is for the machine to read and process a machine-sensible specification of the semantics of the information.  This is a robust and very challenging approach, and largely beyond the current state of the art. A much simpler alternative is for the human Web application developers to hardwire the knowledge into the software so that when the machine runs the software, it does the correct thing with the information." \cite{uschold:sem2001}
\end{quote}
Because relationships among resources are only denoted by a URI, many issues arise around the notion of \textit{context}.  Context-senstive algorithms have been developed to deal with problems such as term disambiguation \cite{schema:magnini2003}, naming conflicts \cite{context:tierney2005}, and ontology integration \cite{wache:onto2001,graphont:udrea2005}. The cause of such problems is the fact that statements, by themselves, ignore the situatedness that defines the semantics of such assertions \cite{Floridi2007Web-2.0}. Similar criticism directed towards the issue of semantics has appeared in other specialized literature \cite{know:sowa1999,woods2004,zadeh2002}. Sheth et. al. \cite{sheth:implicit2005} have framed this issue well and have provided a clear distinction between the various levels of meaning on the Semantic Web.\footnote{A similar presentation is also presented in \cite{uschold:sem2001}.}
\begin{itemize}
\item \textit{Implicit} semantics reside within the minds of humans as a collective consensus and as such, are not explicitly recorded in some machine processable medium. 
\item \textit{Formal} semantics are in a machine-readable format in the form of an ontology and are primarily used for human consumption and for machine hardwiring. 
\item \textit{Soft} semantics are extracted from probabilistic and fuzzy reasoning mechanisms supporting degree of membership and certainty.
\end{itemize}

The model proposed in this article primarily falls within the domain of soft semantics. Simply put, the purpose of the model is to supplement a statement with other statements. These other statements, while being part of the RDF graph itself, serve to contextualize the original statement.

\begin{quote}
``Contextualization is a word first used in sociolinguistics to refer to the use of language and discourse to signal relevant aspects of an interactional or communicative situation."\footnote{Wikipedia (\ttt{http://en.wikipedia.org/wiki/Contextualization}).}
\end{quote}
The supplementary statements serve to expose the relevant aspects of the interaction between a subject and an object that are tied by a relationship defined by a predicate. With respect to the example of the statement $(\ttt{lanl:marko}, \ttt{foaf:knows}, \ttt{ucla:apepe})$, supplementary statements help to answer the question: ``What do you mean by `Marko knows Alberto'?". A notion from Ludwig Wittgenstein's theory of ``language games'' can be aptly borrowed: the meaning of a concept is not universal and set in stone, but shaped by ``a complicated network of similarities, overlapping and criss-crossing'' \cite{witten:pi1973}. Following this line of thought, this article purposes a ``dilated" model of an RDF triple. The dilated triple contextualizes the meaning and enhances the expressiveness of assertions on the Semantic Web.

\section{\label{sec:model}The Dilated Triple Model}

A single predicate URI does not provide the appropriate degrees of freedom required when modeling the nuances of an RDF relationship. The model proposed in this article enhances the expressiveness of a triple such that its meaning is considered within a larger context as defined by a graph structure. It thereby provides a machine with the ability to discern the more fine-grained context in which a statement relates its subject and object.

\begin{figure}[h!]
	\centering
		\includegraphics[width=0.35\textwidth]{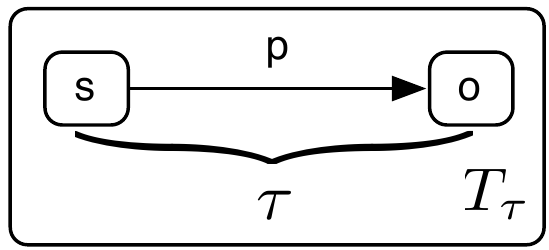}
	\caption{The dilated triple $T_\tau$.}
	\label{fig:dilated-triple}
\end{figure}

In the proposed model, every triple in an RDF graph is supplemented with other triples from the same RDF graph. The triple and its supplements form what is called a \textit{dilated triple}.\footnote{The Oxford English dictionary provides two definitions for the word ``dilate":  ``to expand" and  ``to speak or write at length". It will become clear through the remainder of this article that both definitions suffice to succinctly summarize the presented model.}
\begin{definition}[A Dilated Triple]
Given a set of triples $R$ and a triple $\tau \in R$, a dilation of $\tau$ is a set of triples $T_\tau \subset R$ such that $\tau \in T_\tau$.
\end{definition}
The dilated form of $\tau \in R$ is $T_\tau$. Informally, $T_\tau$ servers to elaborate the meaning of $\tau$. Formally, $T_\tau$ is a graph that at minimum contains only $\tau$ and at maximum contains all triples in $R$. The set of all non-$\tau$ triples in $T_\tau$ (i.e.~$T_\tau \setminus \tau$) are called \textit{supplementary triples} as they serve to contextualize, or supplement, the meaning of $\tau$. Finally, it is worth noting that every supplemental triple in $T_\tau$ has an associated dilated form, so that $T_\tau$ can be considered a set of nested sets.\footnote{The set of all dilated triples forms a \textit{dilated graph} denoted $\mca{T} = \bigcup_{\tau \in R} \{ T_\tau \}$.} An instance of $\tau$, its subject $s$, predicate $p$, object $o$, and its dilated form $T_\tau$, are diagrammed in Figure \ref{fig:dilated-triple}.

A dilated triple can be conveniently represented in RDF using a named graph \cite{named:carroll2005}. Statements using the named graph construct are not triples, but instead, are quads with the fourth component being denoted by a URI or blank node. Formally, $\tau = (s,p,o,g)$ and $g \in U \cup B$. The fourth component is considered the ``graph" in which the triple is contained. Thus, multiple quads with the same fourth element are considered different triples in the same graph. Named graphs were developed as a more compact (in terms of space) way to reify a triple. The reification of a triple was originally presented in the specification of RDF with the \ttt{rdf:Statement} construct \cite{rdfcon:klyne2004}. RDF reification has historically been used to add specific metadata to a triple, such as provenance, pedigree, privacy, and copyright information. In this article, the purpose of reifying a triple is to supplement its meaning with those of additional triples. While it is possible to make additional statements about the dilated triple (i.e.~the named graph component $g$), the motivation behind the dilated triple is to encapsulate many triples within a single graph, not to make statements about the graph \textit{per se}.

The following sections will further explain the way in which a dilated triple contextualizes the meaning of a statement. \S \ref{sec:structure} demonstrates, by means of an example, how supplementary triples augment the meaning of a relationship between two resources. \S \ref{sec:process} discusses how dilated triples can be compared and used by a machine to discern context.

\section{\label{sec:structure}Contextualizing a Relationship}

\begin{figure}[h!]
	\centering
		\includegraphics[width=0.55\textwidth]{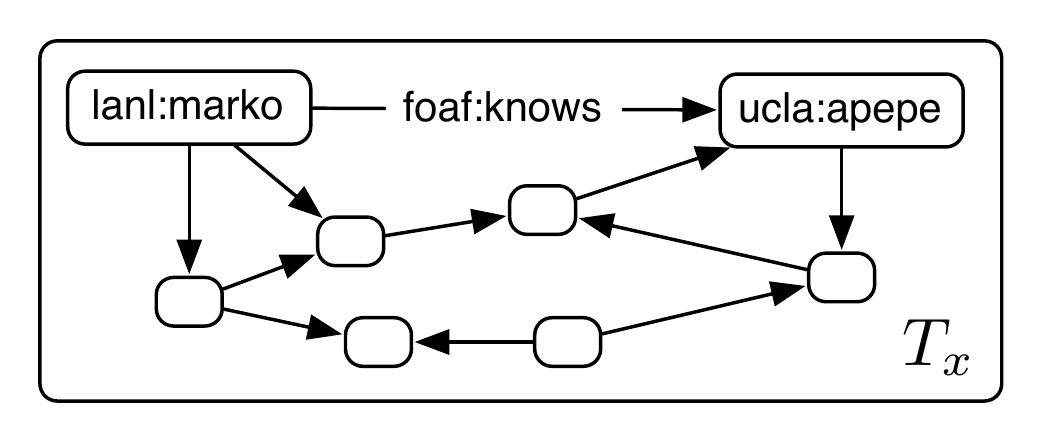}
	\caption{The dilated form of $(\ttt{lanl:marko}, \ttt{foaf:knows}, \ttt{ucla:apepe})$.}
	\label{fig:dilated-marko-alberto}
\end{figure}

The dilated form of $x \in R$, denoted $T_x$, provides a knowledge structure that is suited to contextualizing the meaning of an assertion made about the world. For example, consider the asserted triple
\begin{equation}
	x = (\ttt{lanl:marko}, \ttt{foaf:knows}, \ttt{ucla:apepe}).
\end{equation}
What is the meaning of \ttt{foaf:knows} in this context? For the human, the meaning is made explicit in the specification document of the FOAF (Friend of a Friend) ontology (\ttt{http://xmlns.com/foaf/spec/}), which states:

\begin{quote}
``We take a broad view of `knows', but do require some form of reciprocated interaction (i.e.~stalkers need not apply). Since social attitudes and conventions on this topic vary greatly between communities, counties and cultures, it is not appropriate for FOAF to be overly-specific here."
\end{quote}
Unfortunately, the supplementary information that defines the meaning of \ttt{foaf:knows} is not encoded with the URI itself (nor in the greater RDF graph) and it is only through some external medium (the FOAF specification document) that the meaning of \ttt{foaf:knows} is made clear. Thus, such semantics are entirely informal \cite{uschold:sem2001}. However, even if the complexity of the meaning of \ttt{foaf:knows} could be conveyed by an RDF graph, the nuances of ``knowing" are subtle, such that no two people know each other in quite the same way. In fact, only at the most abstract level of analysis is the relationship of ``knowing" the same between any two people. In order for a human or machine to understand the way in which \ttt{lanl:marko} and \ttt{ucla:apepe} know each other, the complexities of this relationship must be stated. In other words,  it is important to state ``a constraint on the possible ways the world might be" \cite{rdfsem:hayes2004} even if that constraint is a complex graph of relationships. For example, the meaning of \ttt{foaf:knows} when applied to \ttt{lanl:marko} and \ttt{ucla:apepe} can be more eloquently stated as:

\begin{quote}
``Marko and Alberto first met at the European Organization for Nuclear Research (CERN) at the Open Archives Initiative Conference (OAI) in 2005. Six months later, Alberto began a summer internship at the Los Alamos National Laboratory (LANL) where he worked under Herbert Van de Sompel on the Digital Library Research and Prototyping Team. Marko, at the time, was also working under Herbert Van de Sompel. Unbeknownst to Herbert, Marko and Alberto analyzed a scholarly data set that Alberto had acquired at the Center for Embedded Networked Sensing (CENS) at the University of California at Los Angeles (UCLA). The results of their analysis ultimately led to the publication of an article \cite{onthe:rodriguez2008} in Leo Egghe's Journal of Informetrics. Marko and Alberto were excited to publish in Leo Egghe's journal after meeting him at the Institute for Pure and Applied Mathematics (IPAM) at UCLA."
\end{quote}
The facts above, when represented as an RDF graph with triples relating such concepts as \ttt{lanl:marko}, \ttt{ucla:apepe}, \ttt{lanl:herbertv}, \ttt{cern:cern}, \ttt{ucla:ipam}, \ttt{elsevier:joi}, \ttt{doi:10.1016/j.joi.2008.04.002}, etc., serve to form the dilated triple $T_x$. In this way, the meaning of the asserted triple  $(\ttt{lanl:marko}, \ttt{foaf:knows}, \ttt{ucla:apepe})$ is presented in the broader context $T_x$. In other words, $T_x$ helps to elucidate the way in which Marko knows Alberto. Figure \ref{fig:dilated-marko-alberto} depicts $T_x$, where the unlabeled resources and relationships represent the URIs from the previous representation.\footnote{For the sake of diagram clarity, the supplemented triples are unlabeled in Figure \ref{fig:dilated-marko-alberto}. However, please be aware that the unlabeled resources are in fact the URI encoding of the aforementioned natural language example explaining how Marko knows Alberto.}

Even after all the aforementioned facts about Marko and Alberto's ``knowing" relationship are encoded in $T_x$, still more information is required to fully understand what is meant by Marko ``knowing" Alberto. What is the nature of the scholarly data set that they analyzed? Who did what for the analysis? Did they ever socialize outside of work when Alberto was visiting LANL? What was the conversation that they had with Leo Egghe like? Can their \ttt{foaf:knows} relationship ever be fully understood? Only an infinite recursion into their histories, experiences, and subjective worlds could reveal the ``true" meaning of  $(\ttt{lanl:marko}, \ttt{foaf:knows}, \ttt{ucla:apepe})$. Only when $T_\tau = R$,\footnote{For the purpose of this part of the argument, $R$ is assumed to be a theoretical graph instance that includes all statements about the world.} that is, when their relationship is placed within the broader context of the world as a whole, does the complete picture emerge. However, with respect to those triples that provide the most context, a $|T_\tau| \ll |R|$ suffices to expose the more essential aspects of $(\ttt{lanl:marko}, \ttt{foaf:knows}, \ttt{ucla:apepe})$.\footnote{A fuzzy set is perhaps the best representation of a dilated triple \cite{zadeh:fuzzy1965}. In such cases, a membership function $\mu_{T_\tau}: R \rar [0,1]$ would define the degree to which every triple in $R$ is in $T_\tau$. However, for the sake of simplicity and to present the proposed model within the constructs of the popular named graph formalism, $T_\tau$ is considered a classical set. Moreover, a fuzzy logic representation requires an associated membership valued in $[0,1]$ which then requires further statement reification in order to add such metadata. With classic bivalent logic, $\{0,1\}$ iss captured by the membership or non-membership of the statement in $T_\tau$.}\footnote{The choices made in the creation of a dilated triple are determined at the knowledge-level \cite{know:newell1982}. The presentation here does not suppose the means of creation, only the underlying representation and utilization of such a representation.}

\begin{figure}[h!]
	\centering
		\includegraphics[width=0.9\textwidth]{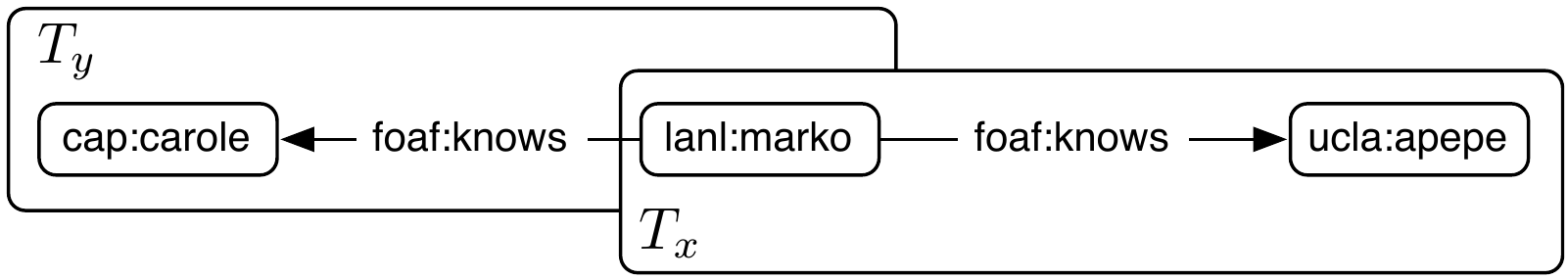}
	\caption{The dilated triples $T_x$ and $T_y$.}
	\label{fig:contextual-predicate}
\end{figure}

Defining a triple in terms of a larger conceptual structure exposes more degrees of freedom when representing the uniqueness of a relationship. For example, suppose the two dilated triples $T_x$ and $T_y$ diagrammed in Figure \ref{fig:contextual-predicate}, where 
\begin{equation}
	x = (\ttt{lanl:marko}, \ttt{foaf:knows}, \ttt{ucla:apepe})
\end{equation}
and
\begin{equation}
	y = (\ttt{lanl:marko}, \ttt{foaf:knows}, \ttt{cap:carole}).
\end{equation}

Both $T_x$ and $T_y$ share the same predicate \ttt{foaf:knows}. However, what is meant by Marko knowing Alberto is much different than what is meant by Marko knowing his mother Carole (\ttt{cap:carole}). While, broadly speaking, it is true that Marko knows both Alberto and Carole, the context in which Marko knows Alberto is much different than the context in which Marko knows Carole. The supplementary triples that compose $T_y$ may be the RDF expression of:

\begin{quote}
``Marko was born in Fairfield, California on November 30$^\text{th}$, 1979. Carole is Marko's mother. Marko's family lived in Riverside (California), Peachtree City (Georgia), Panama City (Panama), and Fairfax (Virginia). During his $10^\text{th}$ grade high-school term, Marko moved with his family back to Fairfield, California."
\end{quote}

It is obvious from these two examples that \ttt{foaf:knows} can not sufficiently express the subtleties that exist between two people. People know each other in many different ways. There are family relationships, business relationships, scholarly relationships, and so on. It is true that these subtleties can be exposed when performing a deeper analysis of the graph surrounding a \ttt{foaf:knows} relationship as other paths will emerge that exist between people (e.g.~vacation paths, transaction paths, coauthorship paths, etc.). The purpose of a dilated triple is to contain these corroborating statements within the relationship itself. The purpose of $T_x$ is to identify those aspects of Marko and Alberto's ``knowing" relationship that make it unique (that provide it the most meaning). Similarly, the purpose of $T_y$ is to provide a finer representation of the context in which Marko knows his mother. The supplementary triples of $T_x$ and $T_y$ augment the meaning of \ttt{foaf:knows} and frame each respective triple $x$ and $y$ in a broader context.\footnote{Examples of other predicates beyond \ttt{foaf:knows} also exist. For instance, suppose the predicates \ttt{foaf:member} and \ttt{foaf:fundedBy}. In what way is that individual a member of that group and how is that individual funded?}

\section{\label{sec:process}Comparing Contexts}

The ``Marko knows" examples from the previous section are reused in this section to explain how dilated triples can assist a machine in discerning and comparing the broader meaning of a statement. In order to present this example, the notion of a contextualized process is introduced. A contextualized process, as defined here, is a human or machine that maintains a perspective or expectation of how the world must be. 

\begin{quote}
``Perspective in theory of cognition is the choice of a context or a reference (or the result of this choice) from which to sense, categorize, measure or codify experience, cohesively forming a coherent belief, typically for comparing with another."\footnote{Wikipedia \ttt{http://en.wikipedia.org/wiki/Perspective\_(cognitive)}.}
\end{quote}
A perspective can be expressed in RDF by simply associating some process with a knowledge-base that defines what that process knows about the world. This knowledge-base can be derived, for example, from the process' history and thus defined as a subgraph of the full RDF graph. With respect to a perspective based on history, it makes sense that a process does not ``experience" the whole RDF graph in a single moment, but instead, traverses through a graph by querying for future states and formalizing a model of past experiences \cite{heylighen:ci1999}. The algorithm by which a process derives a new state is perhaps ``hardwired"  \cite{uschold:sem2001}, but what is returned by that query is dependent upon the structure of the full graph and the process' historic subgraph. Furthermore, by capitalizing on the notion of a contextualization of both the perspective of the process and the meaning of a statement, that query can yield different results for different processes. In general, it is the interaction between the structure of the world and the context of the process that determines process' subjective realization of the world into the future.

Suppose a process were to take a path through an RDF graph that included such concepts as Marko's publications, the members of the Center for Nonlinear Studies, his collaborations with the Center for Embedded Networked Sensing, and the various conferences he has attended. Given this set of experiences, the process has built up a mental model of Marko that primarily includes those aspects of his life that are scholarly.\footnote{It is noted that Marko is a complex concept and includes not only his academic life, but also his personal, business, hobby, etc. lives.} More generally, the process has been pigeonholed into a scholarly ``frame of mind". Without any context, if that process were to inquire about the people that Marko knows (e.g.~$(\ttt{lanl:marko}, \ttt{foaf:knows}, ?o)$), it would learn that Marko knows both Alberto and Carole. However, given the context of the process (i.e.~the history as represented by its traversed subgraph), it will interpret ``knowing" with respect to those people that Marko knows in a scholastic sense. Given Alberto and Carole, Marko knows Alberto in a scholarly manner, which is not true of Carole. There is little to nothing in Marko and Carole's $T_y$ that makes reference to anything scholarly. However, in Marko and Alberto's $T_x$, the whole premise of their relationship is scholarly.

\begin{figure}[h!]
	\centering
		\includegraphics[width=0.9\textwidth]{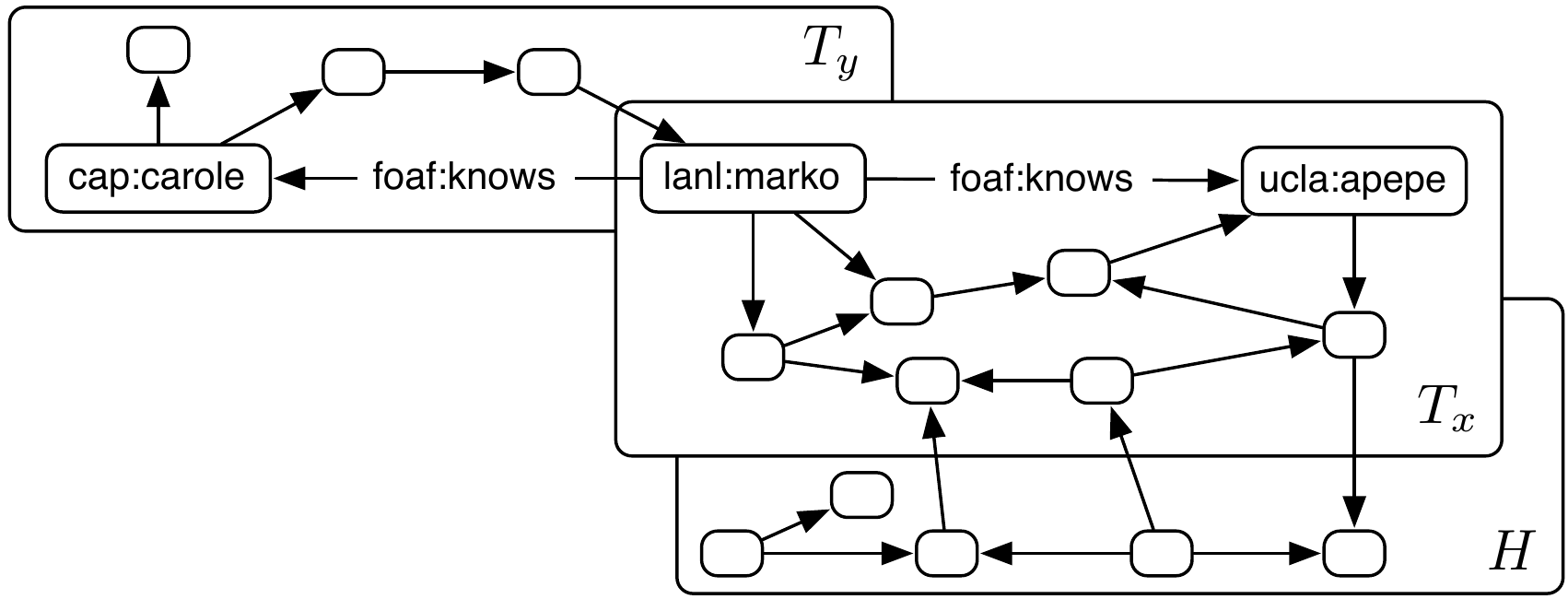}
	\caption{The relationship between the context of a process and a dilated triple.}
	\label{fig:contextual-process}
\end{figure}

A simple way in which the process can make a distinction between the various interpretations of \ttt{foaf:knows} is to intersect its history with the context of the relationships. In other words, the process can compare its history subgraph with the subgraph that constitutes a dilated triple. If $H \subseteq R$ is a graph defining the history of the process which includes the process' traversal through the scholarly aspects of Marko, then it is the case that $| H \cap T_x | > | H \cap T_y |$ as the process' scholarly perspective is more related to Marko and Alberto than it is to Marko and Carole. That is, the process' history $H$ has more triples in common with $T_x$ than with $T_y$. Thus, what the process means by \ttt{foaf:knows} is a ``scholarly" \ttt{foaf:knows}. This idea is diagrammed in Figure \ref{fig:contextual-process}, where $H$ has more in common with $T_x$ than with $T_y$, thus an intersection of these sets would yield a solution to the query $(\ttt{lanl:marko}, \ttt{foaf:knows}, ?o)$ that included Alberto and not Carole.\footnote{$H$ need not be a dynamic context that is generated as a process moves through an RDF graph. $H$ can also be seen as a static, hardwired ``expectation" of what the process should perceive. For instance, $H$ could include ontological triples and known instance triples. In such cases, querying for such relationships as \ttt{foaf:knows}, \ttt{foaf:fundedBy}, \ttt{foaf:memberOf}, etc. would yield results related to $H$ -- biasing the results towards those relationships that are most representative of the process' expectations.} In other words, the history of the process ``blinds" the process in favor of interpreting its place in the graph from the scholarly angle.\footnote{This notion is sometimes regarded as a ``reality tunnel" \cite{nerosoc:wilson1979,prome:wilson1983}.}

The trivial intersection method of identifying the degree of similarity between two graph structures can be extended.  Other algorithms, such as those based on a spreading activation within a semantic graph \cite{spread:collins1975,inform:cohen1987,search:crestani2000,grammar:rodriguez2008} can be used as a more fuzzy and probabilistic means of determining the relative ``semantic distance" between two graphs \cite{semdist:delugach1993}. Spreading activation methods are more analogous to the connectionist paradigm of cognitive science than the symbolic methods of artificial intelligence research \cite{rumelhart:conn1993}. The purpose of a spreading activation algorithm is to determine which resources in a semantic graph are most related to some other set of resources. In general, a spreading activation algorithm diffuses an energy distribution over a graph starting from a set of resources and proceeding until a predetermined number of steps have been taken or the energy decays to some $\epsilon \approx 0$.\footnote{In many ways this is analagous to finding the primary eigenvector of the graph using the power method. However, the energy vector at time step $1$ only has values for the source resources, the energy vector is decayed on each iteration, and finally, only so many iterations are executed as a steady state distribution is not desired.} Those resources that received the most energy flow during the spreading activation process are considered the most similar to the set of source resources. With respect to the particular example at hand, the energy diffusion would start at the resources in $H$ and the results would be compared with resources of $T_x$ and $T_y$. If the set of resource in $T_x$ received more energy than those in $T_y$, then the dilated triple $T_x$ is considered more representative of the context of $H$.\footnote{Spreading activation on a semantic graph is complicated as edges have labels. A framework that makes use of this fact to perform arbitrary path traversals through a semantic graph is presented in \cite{grammar:rodriguez2008}.}

By taking advantage of the supplementary information contained within a dilated triple, a process has more information on which to base its interpretation of the meaning of a triple. To the process, a triple is not simply a string of three symbols, but instead is a larger knowledge structure which encapsulates the uniqueness of the relationship. The process can use this information to bias its traversal of the graph and thus, how it goes about discovering information in the graph.

\section{Conclusion}

While a single word may be used in any number of contexts, its precise meaning is never entirely the same \cite{witten:pi1973}. This is not a difficult or unusual problem for human beings, as meaning is naturally contextualized by other facts about the world. Thus, humans can speak in broad terms without loss of meaning, as the finer details of the situation are understood in context. Humans possess an awareness of their current situation and a set of expectations about the world which allow them to discern the subtleties of its structure. In an effort to bridge the gap between the natural context sensitivity of human communication and the rigid context independence of the Semantic Web, this article has presented a simple model in which the meaning of a statement is defined in terms of other statements, much like the words of a dictionary are defined in terms of each other. This model serves to strengthen the association between concepts and the contexts in which they are to be understood.

   \bibliographystyle{spmpsci}
   \bibliography{../marko}

\end{document}